\pdfoutput=1
\documentclass[sigconf]{acmart}

\AtBeginDocument{%
  }

\acmConference[MM '23]{Proceedings of the 31th ACM International Conference on Multimedia}{October 29--November 3,
  2023}{Ottawa, Canada}
\copyrightyear{2023}
\acmYear{2023}
\setcopyright{acmlicensed}\acmConference[MM '23]{Proceedings of the 31st ACM International Conference on Multimedia}{October 29-November 3, 2023}{Ottawa, ON, Canada}
\acmBooktitle{Proceedings of the 31st ACM International Conference on Multimedia (MM '23), October 29-November 3, 2023, Ottawa, ON, Canada} \acmPrice{15.00}
\acmDOI{10.1145/3581783.3611930} \acmISBN{979-8-4007-0108-5/23/10}

\acmSubmissionID{1037}

\usepackage{float}
\usepackage{stfloats}
\usepackage{graphicx}
\usepackage{hyperref}

\begin{document}
\begin{sloppypar}

\title{AutoPoster: A Highly Automatic and Content-aware Design System for Advertising Poster Generation}

\author{Jinpeng Lin}
\authornote{Both authors contributed equally to this research.}
\author{Min Zhou}
\authornotemark[1]
\affiliation{%
  \institution{Alibaba Group}
  \city{Beijing}
  \country{China}
}
\email{jplinforever@gmail.com}
\email{yunqi.zm@alibaba-inc.com}


\author{Ye Ma}
\affiliation{%
  \institution{Alibaba Group}
  \city{Beijing}
  \country{China}
}
\email{maye.my@alibaba-inc.com}

\author{Yifan Gao}
\affiliation{%
 \institution{University of Science and Technology
of China}
\city{Hefei}
  \country{China}}
\email{eafn@mail.ustc.edu.cn}

\author{Chenxi Fei}
\affiliation{%
  \institution{Alibaba Group}
  \city{Hangzhou}
  \country{China}}
\email{corey.fcx@alibaba-inc.com}

\author{Yangjian Chen}
\author{Zhang Yu}
\affiliation{%
  \institution{Alibaba Group}
  \city{Hangzhou}
  \country{China}}
\email{yicai.cyj@alibaba-inc.com}
\email{zy99945@alibaba-inc.com}

\author{Tiezheng Ge}
\authornote{Corresponding author}
\affiliation{%
  \institution{Alibaba Group}
  \city{Beijing}
  \country{China}}
\email{tiezheng.gtz@alibaba-inc.com}

\renewcommand{\shortauthors}{Jinpeng Lin et al.}


\begin{abstract}
  Advertising posters, a form of information presentation, combine visual and linguistic modalities. Creating a poster involves multiple steps and necessitates design experience and creativity. This paper introduces AutoPoster, a highly automatic and content-aware system for generating advertising posters. With only product images and titles as inputs, AutoPoster can automatically produce posters of varying sizes through four key stages: image cleaning and retargeting, layout generation, tagline generation, and style attribute prediction. To ensure visual harmony of posters, two content-aware models are incorporated for layout and tagline generation. Moreover, we propose a novel multi-task Style Attribute Predictor (SAP) to jointly predict visual style attributes. Meanwhile, to our knowledge, we propose the first poster generation dataset that includes visual attribute annotations for over 76k posters. Qualitative and quantitative outcomes from user studies and experiments substantiate the efficacy of our system and the aesthetic superiority of the generated posters compared to other poster generation methods.
\end{abstract}




\maketitle

\section{Introduction}\label{intro}

\begin{figure}
  \vspace{0.5cm}
  \includegraphics[width=0.5\textwidth]{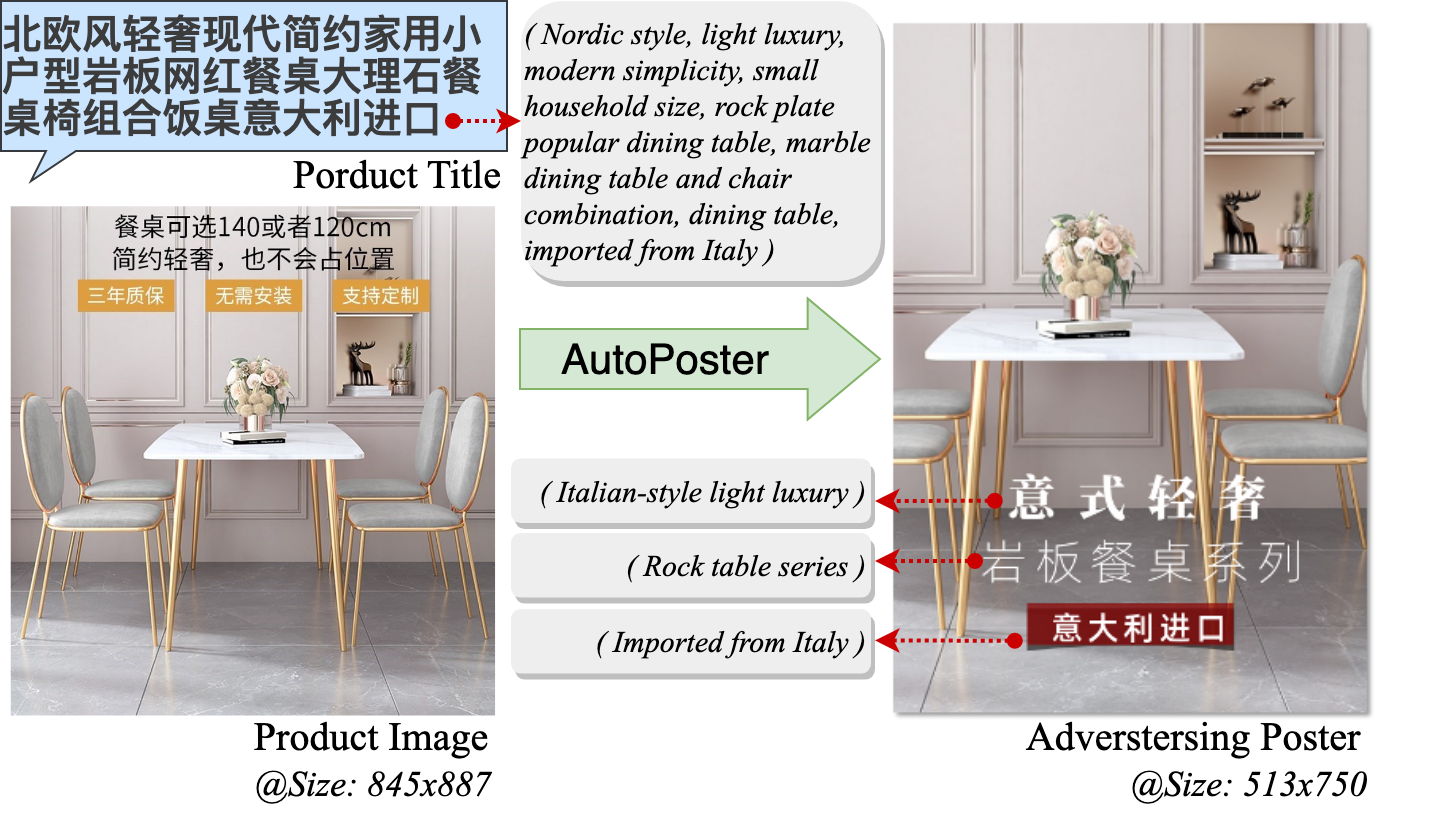}
   \vspace{-18pt}
  \caption{Given a product image, product title, and target poster size, our AutoPoster method can automatically generate the corresponding poster. To enhance readability, English translations of the Chinese taglines are \textit{italicized}, and subsequent figures are treated in the same manner.}
  \label{fig:teaser}
  \vspace{-10pt}
\end{figure}

Conveying comprehensive product information is crucial for e-commerce platforms, especially when potential customers are browsing without a clear purchase objective. Advertising posters, which incorporate product images and diverse graphic design elements such as taglines, underlays, and logos, effectively display product appearances, brands, and unique selling points. 

With multiple elements of different modalities to consider, poster creation can be a challenging task that involves three main aspects and requirements. First, the layout of images and various graphic elements must meet the target size requirements and be harmonious, while highlighting the main subject. Second, concise and structured taglines must be generated based on the product information to quickly convey messages to customers. Third, to achieve a harmonious and visually appealing poster, all elements within the poster are interconnected. Therefore, manual poster creation often requires a significant amount of time and creativity, particularly when creating posters of different sizes for various products. While some automated methods~\cite{DBLP:journals/corr/abs-1908-10139, DBLP:conf/chi/GuoJSLL0C21, DBLP:conf/chi/VaddamanuAGSC22, yang2016automatic} are available to assist with poster creation, they often lack sufficient automation and require users to prepare target-sized images and tagline content in advance. Additionally, they depend heavily on manual rules to ensure visual effects, resulting in reduced flexibility and diversity.

\begin{figure*}[ht!]
  \centering
  \includegraphics[width=0.97\textwidth]{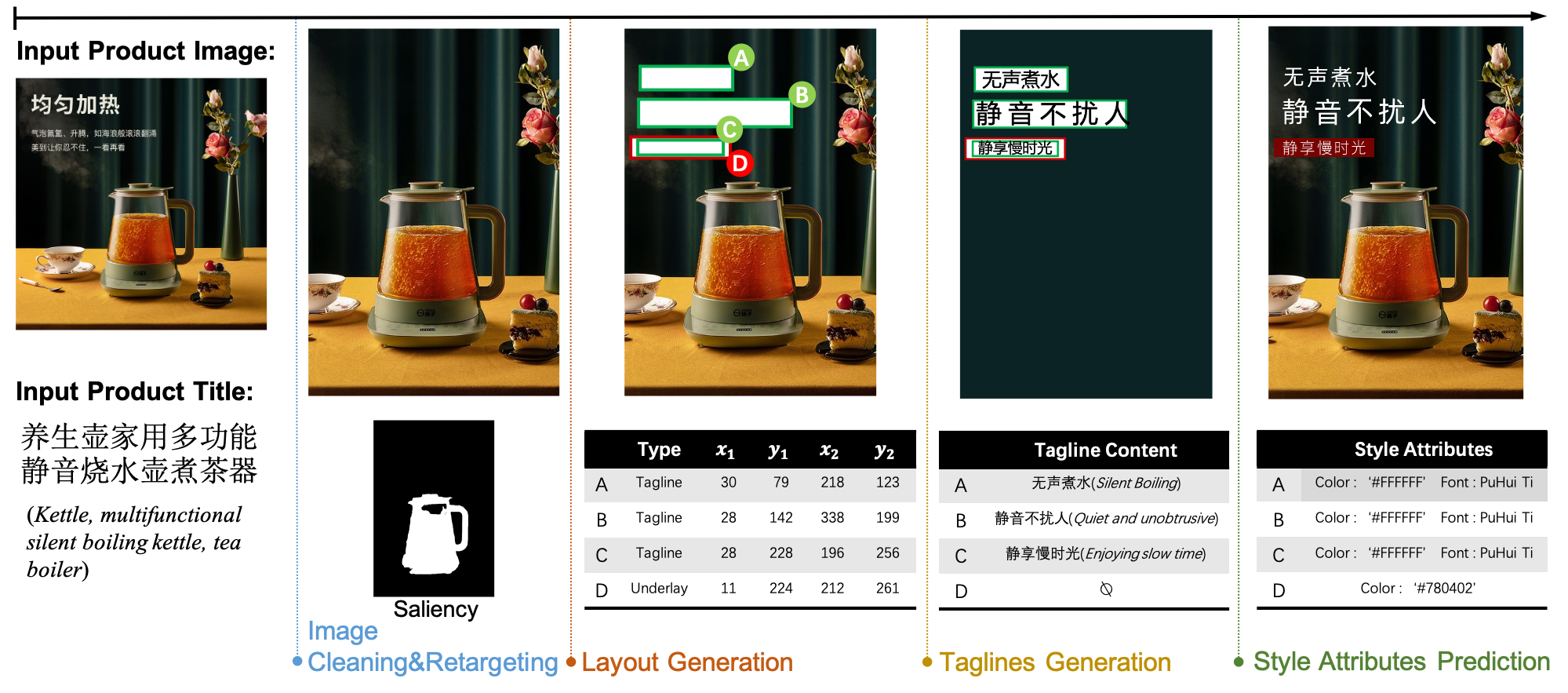}
  \vspace{-8pt}
  \caption{AutoPoster creates posters by following a four-stage process that utilizes product images and titles. The stages are displayed in a sequential manner from left to right.}
  \label{pipeline}
  \vspace{-5pt}
\end{figure*}

In this paper, we propose a novel and highly automated method for generating advertising posters, called AutoPoster. As shown in Fig.~\ref{fig:teaser}, an arbitrary product image and product description are all the information provided by the user to generate a complete poster of a specified size, and the user can continue to adjust the layout, tagline content and style attributes of the graphic elements. Professional designers often first organize visual elements (e.g. product image) and then prepare graphic elements (e.g. taglines, logos) accordingly. Besides, to achieve a harmonious and visually appealing poster, all elements within the poster are interconnected. Guided by these principles, AutoPoster can be divided into four steps:  1) image cleaning and retargeting. Assisted with detection, inpainting, saliency, and outpainting models, graphic elements on the product image are erased, and the image is retargeted to the target poster size while maintaining the subject unchanged. 2) layout generation. We utilize CGL-GAN~\cite{DBLP:conf/ijcai/ZhouXMGJX22} and ICVT~\cite{cao2022geometry} to arrange the number and position of graphic elements (text, logo, and underlay) according to image contents. 3) tagline generation. A multi-modal generative model~\cite{DBLP:journals/corr/abs-2204-12974} is used to yield tagline content based on image content, product information, and layouts. 4) style attribute prediction. We introduce an innovative Style Attribute Predictor~(SAP), a non-autoregressive and multi-task transformer that models the visual relationship between images and graphic elements, as well as that within graphic elements. In other words, SAP can simultaneously predict the font typography, dominant color, stroke color, and gradient color of taglines and underlays according to image contents and graphic layouts. 

We trained several models using a dataset of 76,960 annotated posters, which is the first large-scale dataset for poster generation. The annotation encompasses tagline content, graphic element position, and style attributes. To annotate font typography that is difficult for humans, we trained a font recognition model with self-supervised learning. Qualitative and quantitative experiments show the effectiveness and superiority of AutoPoster and SAP.

The contributions of this paper can be summarized as follows:
 \begin{itemize}
     \item We propose a novel approach to model poster design through four key steps: image cleaning and retargeting, layout generation, tagline generation, and style attribute prediction. Our method requires only product images, information such as product titles, and desired poster dimensions from users to produce complete posters.
    
     \item For style attribute prediction, we introduce a novel multi-task and non-autoregressive multi-modal sequence model that learns the relationships among graphic element attributes, their positions, and image content.
     
     \item To highlight and maintain subjects in poster design, we utilize multiple models to clean product images and retarget them to the target sizes.
     
     \item To our knowledge, we constructed the first large poster generation dataset, which contains annotations of tagline content and various visual attributes of graphic elements. And we train a self-supervised model to label the font typography.

 \end{itemize}

\section{Related Work}
The development of an information representation system encompasses various interdisciplinary tasks and topics. We categorize the related work from the comprehensive automatic design generation system down to the individual phases.

\paragraph{\textbf{Automatic information presentation generation system.}} To save manual effort, various automatic methods~\cite{snippet, DBLP:journals/jcst/QiangFYGZS19, Wang_2022_CVPR, DBLP:conf/icassp/JinXSL22} for information presentation have been developed. Two similar works are Vempati \textit{et al.}~\cite{vempati2020enabling}  and Vinci~\cite{DBLP:conf/chi/GuoJSLL0C21}, which can automatically generate posters based on product images and taglines. AutoPoster differs in three main ways: a) it can generate more flexible layouts and any number of taglines on the image based on the product title; b) it can handle a wider range of product images, including images with texts and photo-style images (naturally photographed product images); c) it is equipped with an image-aware style attribute prediction model and can automatically predict more fine-grained attributes, such as font and stroke color. While Vinci and Vempati \textit{et al.}~\cite{vempati2020enabling} only adjust the text color by the contrast between texts and backgrounds, ensuring readability but ignoring color harmonization.

\begin{figure*}[ht!]
  \centering
  \includegraphics[width=0.97\linewidth]{{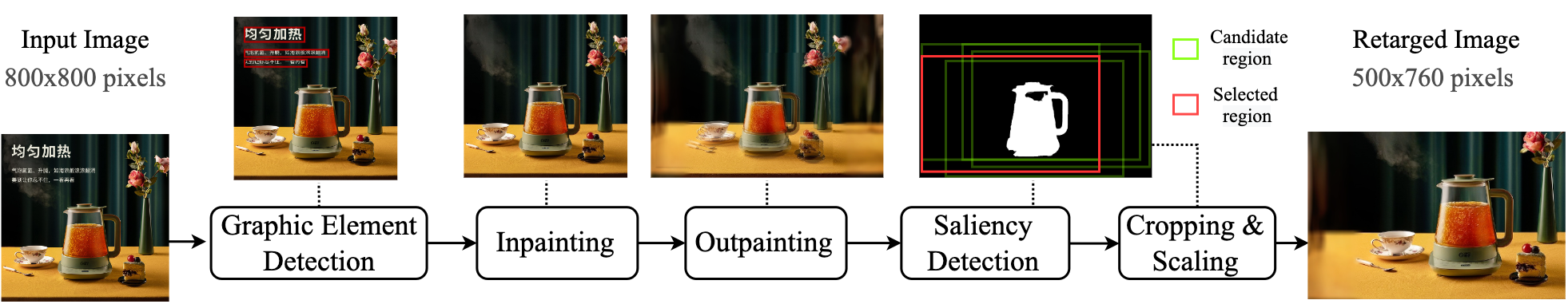}}
  \vspace{-7pt}
  \caption{Five key steps for image cleaning and retargeting pipeline.}
  \Description{Image retargeting pipeline, }
  \label{retargeting}
\end{figure*}

\paragraph{\textbf{Image retargeting.}}
Image retargeting means changing image sizes while keeping the visual content as much as possible. The most straightforward algorithm is cropping~\cite{vempati2020enabling, snippet}. However, when the aspect ratios of the product image size and target poster size are quite different, these methods will inevitably crop off parts of subjects, affecting the product presentation. Warping-based methods~\cite{DBLP:journals/tog/AvidanS07,DBLP:conf/iccv/ChoPOTK17} armed with deep learning can avoid this problem. However, they aim at retaining the whole visual content and may alter the product regions, which is also an unwanted situation in poster design. In contrast to these methods, we propose a combination of cropping and outpainting as a way to tackle the above problems.

\paragraph{\textbf{Layout generation.}}
For graphic layout generation, previous works~\cite{DBLP:journals/tog/JacobsLSBS03,DBLP:conf/iui/SchrierDJWS08,DBLP:conf/chi/KumarTAK11} usually utilize templates or heuristic methods. Recently, CGL-GAN~\cite{DBLP:conf/ijcai/ZhouXMGJX22} and ICVT~\cite{cao2022geometry} utilize transformers\cite{vaswani2017attention} to generate image-content-aware layouts and only require posters with element positions labeled for training. We introduce these two methods to produce layouts conditioned on product images.

\paragraph{\textbf{Tagline generation.}}
Existing automatic poster generation systems~\cite{DBLP:conf/chi/VaddamanuAGSC22,vempati2020enabling,visualtextual} lack the ability to generate taglines, which users need to input manually. Generating taglines automatically on the image would simplify user interactions. The CapOnImage model~\cite{DBLP:journals/corr/abs-2204-12974} produces taglines for various positions on the product image based on visual and textual contextual information. In this work, we introduce the CapOnImage model for tagline generation.

\paragraph{\textbf{Style attribute prediction}}
The style attributes in this paper can be divided into two main categories: font typography and color. For predicting font typographies, Zhao \textit{et al.}~\cite{fontweb} proposes a model to choose a font given a web design. FontMatcher~\cite{FontMatcher} matches a harmonious font to the input image. Visual Font Pairing~\cite{jiang2019visual} focuses on finely recommending suitable paired fonts for a known font. The above methods pick fonts without considering both the image contexts and font selections of other elements. We design a transformer-based model and utilize cross-attention to perceive the image contexts and self-attention for internal element contexts. For color design, some works~\cite{AliJahanian2013RecommendationSF} use color harmonic models~\cite{harmonycolor} for rule-based color selection. ~\cite{visualtextual} adopts the topic-dependent templates in harmonic color design. Leveraging color theory~\cite{jameson1964theory}, ~\cite{DBLP:conf/chi/VaddamanuAGSC22} extract the color palette of the whole image and determine the color considering the contrast ratio to ensure clear visibility. In contrast to these rule-based algorithms, we take a data-driven approach to color prediction similar to the image 
colorization problem~\cite{zhang2016colorful} and extend the font prediction model to a multi-task one.

\section{methodology}

In this section, we begin by defining the design space, and then we provide a detailed description of each module in AutoPoster.

\subsection{Formalized Poster Design Space}\label{Formalized Design Space}
As mentioned earlier, the entire process is divided into four steps illustrated in Fig.~\ref{pipeline}. The image cleaning and retargeting module changes product images to the desired size, maintaining subjects and eliminating existing graphic elements. The layout generation module determines the quantity, variety, and position of graphic elements, while the tagline generation module produces pertinent captions. To guarantee a visually appealing design, the style predictor estimates various visual stylistic attributes. 

\begin{table}[ht!]
\caption{Attributes to be predicted for each type of element.}
\vspace{-7pt}
\label{tab:deisgnattr}
\scalebox{0.73}{
  \begin{tabular}{c|cccc}
    \toprule
    Element Type & Dominate Color & Gradient Color & Stroke Color & Font Typography \\
    \midrule
    Underlay & \checkmark & \checkmark & - & - \\
    Text & \checkmark & \checkmark & \checkmark & \checkmark \\
  \bottomrule
\end{tabular}
}
\end{table}

The corresponding design space can be structured in three dimensions: 
\textbf{Layout design space.}
A layout is an arrangement of $N$ graphic elements $\left\{e_1, e_2, \cdot \cdot \cdot, e_N\right\}$. Each element comprises the category, position, and size, represented as $e_i =[c, x_1, y_1, x_2, y_2]$ ($c \in \left\{background~image, logo, underlay, tagline \right\}$). 
\textbf{Tagline design space.} This paper focuses on creating Chinese taglines. A tagline of length $L$ is denoted as $[{w_1, w_2, \cdot \cdot \cdot, w_L}$], where $w_i$ represents a Chinese character.
\textbf{Visual style attributes design space.} The style attributes of graphic elements here are dominant color $Clr^d$, stroke color $Clr^s$, gradient color $Clr^g$, and font typography $F$. For an element $e_i$, its visual style attributes are formulated as $a_i = \left\{Clr^d_i, Clr^s_i, Clr^g_i, F_i \right\}$. As shown in Table~\ref{tab:deisgnattr}, depending on the category, $e_i$ contains different subsets of $a_i$.

\begin{figure*}[ht!]
  \centering
  \includegraphics[width=0.97\textwidth]{{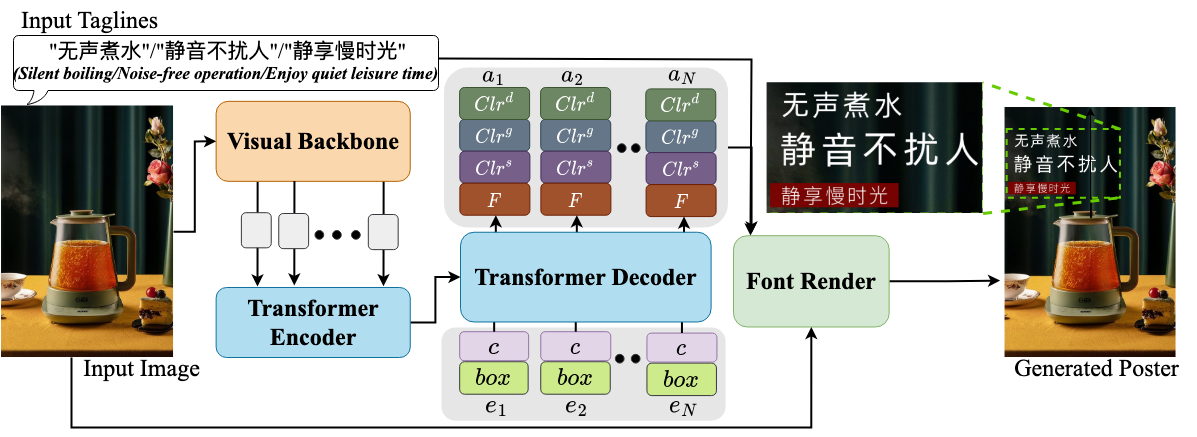}}
  \vspace{-10pt}
  \caption{Model framework of the proposed $SAP$, which takes the image $I$ and layout $\left\{e_1, \dots, e_N\right\}$ as input, and outputs attributes $a_i$ for each element $e_i$. With predicted attributes $\left\{a_1, \dots, a_N \right\}$, layout, taglines and $I$, an advertising poster can be rendered.}
  \label{fig:StyleP}
\end{figure*}

\subsection{Image Cleaning and Retargeting}\label{retargeting_section}
As mentioned in Section~\ref{intro}, there are some rules for the background image of advertising posters: product regions must remain unaltered, surrounding regions can be modified to a greater extent, and the entire image must be visually harmonious and clean. To meet these requirements, we've developed a five-step image cleaning and retargeting pipeline, which is illustrated in Fig.~\ref{retargeting}.

\textbf{Graphic element detection.} An object detection model~\cite{dai2021dynamic} is trained to detect logos, taglines, and underlays in input images. The detected locations are used to create binary masks, indicating areas for inpainting.
\textbf{Inpainting.} We employ an off-the-shelf inpainting model~\cite{Suvorov_2022_WACV} to remove the detected graphic elements and ensure the image is clean for later design processes.
\textbf{Outpainting.} We use a self-supervised approach to train an outpainting model~\cite{DBLP:conf/iccv/KrishnanTSMLBF19}. This model is used to extend image regions seamlessly, avoiding cropping issues due to significant differences in aspect ratios between the target and image. \textbf{Saliency detection.} An off-the-shelf saliency detection model~\cite{DBLP:conf/aaai/WangCZZ0G20} 
 is employed to detect product regions with saliency maps.
\textbf{Cropping and scaling.} Utilizing saliency maps, we initially crop images to achieve the target aspect ratio before scaling them to the desired size. We slide the maximum sub-window with the target aspect ratio across the saliency map to find the region with the highest saliency scores to ensure the product is complete and prominent. We use integral image~\cite{DBLP:conf/cvpr/ViolaJ01} and convolution techniques to speed up this process. Multiple optimal regions may exist, so preferences are set based on the target aspect ratio to determine the final selected region. As Fig.~\ref{retargeting} shows, when the target aspect ratio is greater than one, we find the candidate region with the product center closest to the left or right center of the region. Conversely, for images with aspect ratios less than one, we set the preference at the bottom center.

\subsection{Layout Generation}\label{Layout Generation}
After retargeting the product image to match the poster size, we arrange logos, underlays, and taglines using content-aware layout generation models: CGL-GAN~\cite{DBLP:conf/ijcai/ZhouXMGJX22} and ICVT~\cite{cao2022geometry}. These models utilize product images and saliency maps as input and transformer architecture to learn the relationship between graphic elements and product images. The output is the graphic element layout, denoted as $\{e_1, e_2, \cdot \cdot \cdot, e_N\}$. The training process only requires labeled element positions in posters, not pairs of clean images and posters.

\subsection{Tagline Generation}\label{tagline generation}
Once the layouts have been set, we generate taglines with various styles and content to enhance the appeal of posters. The tagline content may include selling points, click-through guides, benefit points, and more. We follow the approach of~\cite{DBLP:journals/corr/abs-2204-12974} and design an automatic tagline generation module. This module utilizes multi-modal text generation techniques to generate suitable tagline content by taking into account the product image, product title (description information), and layout of all graphic design elements.

\subsection{Style Attribute Prediction}
Rendering graphic elements with style attributes onto the background image is necessary for creating a complete poster, and the rendering quality significantly impacts the readability, information conveyance, and visual aesthetic of the poster. In this section, we present the SAP, which can estimate visual style attributes for all graphic elements based on the product image and layout.

\subsubsection{\textbf{Discrete Representation of Attributes}}
As shown in Table~\ref{tab:deisgnattr}, the style attributes can be summarized as color and font. For model training, we make a uniform discrete representation of these two attributes. Following~\cite{zhang2016colorful}, we perform color-related attributes (dominant color, stroke color, and gradient color) in CIE Lab color space, in which the color distance is more consistent with human perception. Specifically, we quantize the ab space into bins with grid size 10 and only get $313$ values which are in-gamut like~\cite{zhang2016colorful} and evenly quantize the lightness of the LAB color to $10$ values. Therefore, each color can be represented by two discrete values. For the font attribute of taglines, we use 62 commonly used font typography for design, so $F \in [1, 62]$.

\subsubsection{\textbf{Model Architecture}}
Fig.~\ref{fig:StyleP} displays the model overview of the style attribute predictor, $SAP$. It is based on the transformer model\cite{vaswani2017attention} and contains an encoder and a decoder. $SAP$ takes an image $I$ and graphic layout information $E=\{e_i=(x_{1i}, y_{1i}, x_{2i}, y_{2i}, c_i) | i \in [1, N]\}$ as input, and outputs the attributes of all graphic elements $\{ a_i = \left\{Clr^d_i, Clr^s_i, Clr^g_i, F_i \right\} | i \in [1, N]\}$. 

\textbf{Encoder.} A CNN visual backbone is first used for extracting local feature information from product images. More precisely, given an image $I \in \mathbb{R}^{H\times W\times C}$ in RGB space. We first use a ResNet~\cite{resnet} to encode $I$ into a $P\times P$ visual feature with a downscale factor of 16. Then the 2D features are flattened with position embedding and sent to the VIT~\cite{dosovitskiy2020image} transformer to capture long-distance relations. Finally, the encoder generates a visual feature $f_v \in \mathbb{R}^{l\times d}$, where $l=HW/P^2$ is the length of the flattened feature sequence, and $d$ is the embedding dimension.

\vspace{-10pt}
\begin{equation}
 f_v=encoder(I)
    \label{eq:encoder}
\end{equation}

\textbf{Decoder.} The decoder takes the visual feature $f_v$ and graphic layout information $E$ as input. Each block of the decoder consists of multiple layers of self-attention and cross-attention. The self-attention layer models the relationship between different graphic elements, while the cross-attention mechanism queries visual memory features for graphic elements. The decoder predicts the attributes $a_i$ for each element $e_i$, which can be formulated as:

\begin{equation}
\{a_1, \dots, a_N\} =\text{decoder}(f_v, \{e_1, \dots, e_N\})
\label{eq:decoder}
\end{equation}

The attributes of each element can be decoded in a non-autoregressive~\cite{nar} or autoregressive~\cite{seq2seq} manner. We analyze the differences in the subsequent experiments.

\subsubsection{\textbf{Multi-Task Training}}
We adopt a joint optimization strategy for inter-connected style attributes, namely ${Clr^d, Clr^s, Clr^g, F}$. For instance, the selection of dominant color can impact that of stroke color. Moreover, designers may add a stroke to the tagline if the dominant color is insufficient for readability. Thus, we allow $SAP$ to predict multiple attributes concurrently. Our experiments indicate that multi-task joint optimization surpasses single-task learning, thereby revealing the fundamental interconnectedness of these attributes. As style attributes are often long-tail distributed, we utilize focal-loss~\cite{focalloss} to optimize the model. Thus, $SAP$ is trained with the following loss:

\vspace{-10pt}
\begin{equation}
\begin{aligned}
\left\{\hat{a}_1, \dots, \hat{a}_N\right\} = SAP(I, E) 
\end{aligned}
\end{equation}

\vspace{-10pt}
\begin{equation}
\begin{aligned}
Loss = \frac{1}{NK} \sum_{i=1}^N \sum_{j=1}^K \lambda_j FocalLoss(\hat{a}^j_i, a^j_i)
\end{aligned}
\end{equation}

where $\lambda_j$ is used to adjust the loss weight of individual attributes, $K$ denotes the number of attributes, and $\hat{a}^j_i$ represents the predicted $j$-th attribute of the $i$-th element.

\subsubsection{\textbf{Underlay Retrieval}}
To increase the diversity of underlay shapes, we generated a database of various underlays extracted from SVGs designed by professionals. Each underlay is matched with the nearest neighbor from the database according to its size and ratio. The selected underlay is adjusted to fit the predicted position of the underlay element and its color attributes are altered to match the prediction of our $SAP$.

\begin{figure}[ht!]
  \vspace{-5pt}
  \includegraphics[width=0.9\linewidth]
  {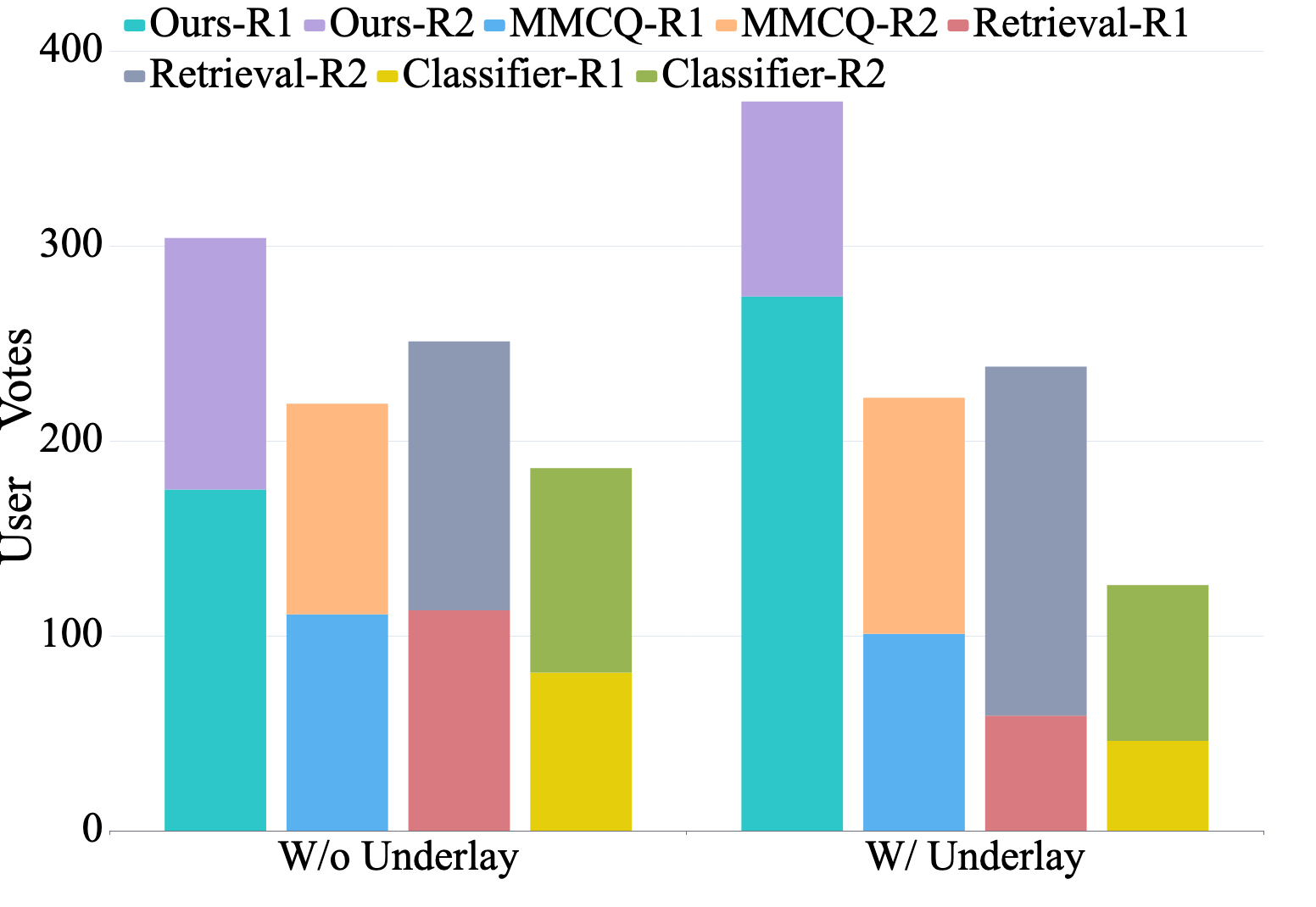}
  \vspace{-15pt}
  \caption{User survey results of ours and other methods on the R1 and R2 scores.}
  \label{fig:userstudy2}
  \vspace{-5pt}
\end{figure}

\vspace{-10pt}
\section{EXPERIMENT}
In this section, we evaluate our proposed method using real-world data. In this section, we evaluate our proposed method using real-world data. We start by introducing the dataset and then demonstrate the effectiveness of $SAP$ through comparisons with other methods and ablation studies. Next, we compare the overall performance of our proposed method with other poster generation approaches and show its ability to generate posters of various sizes.

\subsection{Dataset} 
We obtained 76,960 advertising posters from an e-commerce platform and used 69,249 for training and 7,711 for quantitative tests. Additionally, we collected 46 product images for a user study. The posters are manually designed, providing a diverse set of layouts, taglines, and visual styles across various product categories. We annotated each poster for the layout, tagline content, and color attributes of each graphic design element. For tagline font typography, we trained a font-weight-aware font classifier using a self-supervised approach. Dataset is available at https://tianchi.aliyun.com/dataset/159829.

\begin{table*}
\vspace{-10pt}
  \caption{The ablation study results for SAP. The best results are bolded, and the second-best results are underlined.}
  \label{tab:ablationstudy}
  \vspace{-8pt}
  \scalebox{1.}{
  \begin{tabular}{c|cccccccc}
    \toprule
    Methods & D\text{-}ab (MSE$\downarrow$) & D\text{-}light(MAE$\downarrow$) & G(Acc$\uparrow$) & G\text{-}ab(MSE$\downarrow$) & G\text{-}light(MAE$\downarrow$) & S(Acc$\uparrow$) & S\text{-}ab(MSE$\downarrow$) & S\text{-}light(MAE$\downarrow$)\\
    \midrule
    $Random$                          & 62.71 & 3.90 & 5.0\% & 86 & 2.89 & 16.3\% & 86.3 & 4.48 \\
    $D\text{-}color$                  & 22.81 & \textbf{2.28} & - & - & - & - & - & - \\ 
    $D\text{-}color + S\text{-}color$ & 22.60 & 3.07 & - & - & - & \underline{35.2} \% & \textbf{28.24} & \underline{2.99}  \\ 
    $D\text{-}color + G\text{-}color$ & \underline{22.52} & 2.43 & 27.2\% & \underline{41.50} & \underline{1.78} & - & - & - \\ 
    \midrule
    ours w/AR & 25.64 & 3.0 & \underline{29.0\%} & 46.01 & 1.90 & 28.4\% & 30.37 & 3.23 \\
    ours & \textbf{22.19} & \underline{2.33} & \textbf{30.0\%} & \textbf{41.44} & \textbf{1.78} & \textbf{39.3\%} & \underline{28.40} & \textbf{2.76} \\
  \bottomrule
\end{tabular}
}
\end{table*}

\subsection{Compare SAP with Other Methods}
We compare our SAP model to the following three methods for coloring taglines since there are few works focusing on this task. 

\begin{itemize}
     \item{\textbf{MMCQ}~\cite{DBLP:conf/chi/VaddamanuAGSC22}}: The approach involves extracting the main colors from both local patch and global images separately. Afterward, the color with the highest contrast between extracted colors is selected as the predicted result.
     
     \item{\textbf{Retrieval}}: This method matches tagline colors by using color histograms of tagline regions in images. During inference, it selects the most similar tagline from the library based on the histograms and uses its color as the result.
     \item{\textbf{Classifier}~\cite{fontweb}}: The classifier network estimates the category of tagline color after quantifying the color space.
\end{itemize}

\subsubsection{\textbf{Qualitative Evaluation}} We assess the efficacy of our SAP model via a user survey, which evaluates 30 groups of four posters produced by various methods concerning their readability, harmony, and visual aesthetics. Out of these, 15 groups require color prediction exclusively for taglines (w/o underlays), while the remaining 15 groups involve underlays (w/ underlays), for which our approach predicts the underlay color. Participants select the best (R1) and second-best images (R2) within each group. A higher R1 indicates better outcomes, and R1+R2 reflects greater robustness with fewer inferior cases. The evaluation comprises 30 participants, and the voting results are illustrated in Fig.~\ref{fig:userstudy2}. Our method has the most votes in both R1 and R1+R2, reflecting its superior visual quality. Furthermore, the prediction of underlay color enhances the vote percentage of our method.

\begin{figure}
\vspace{-5pt}
  \centering
  \includegraphics[width=0.9\linewidth]{{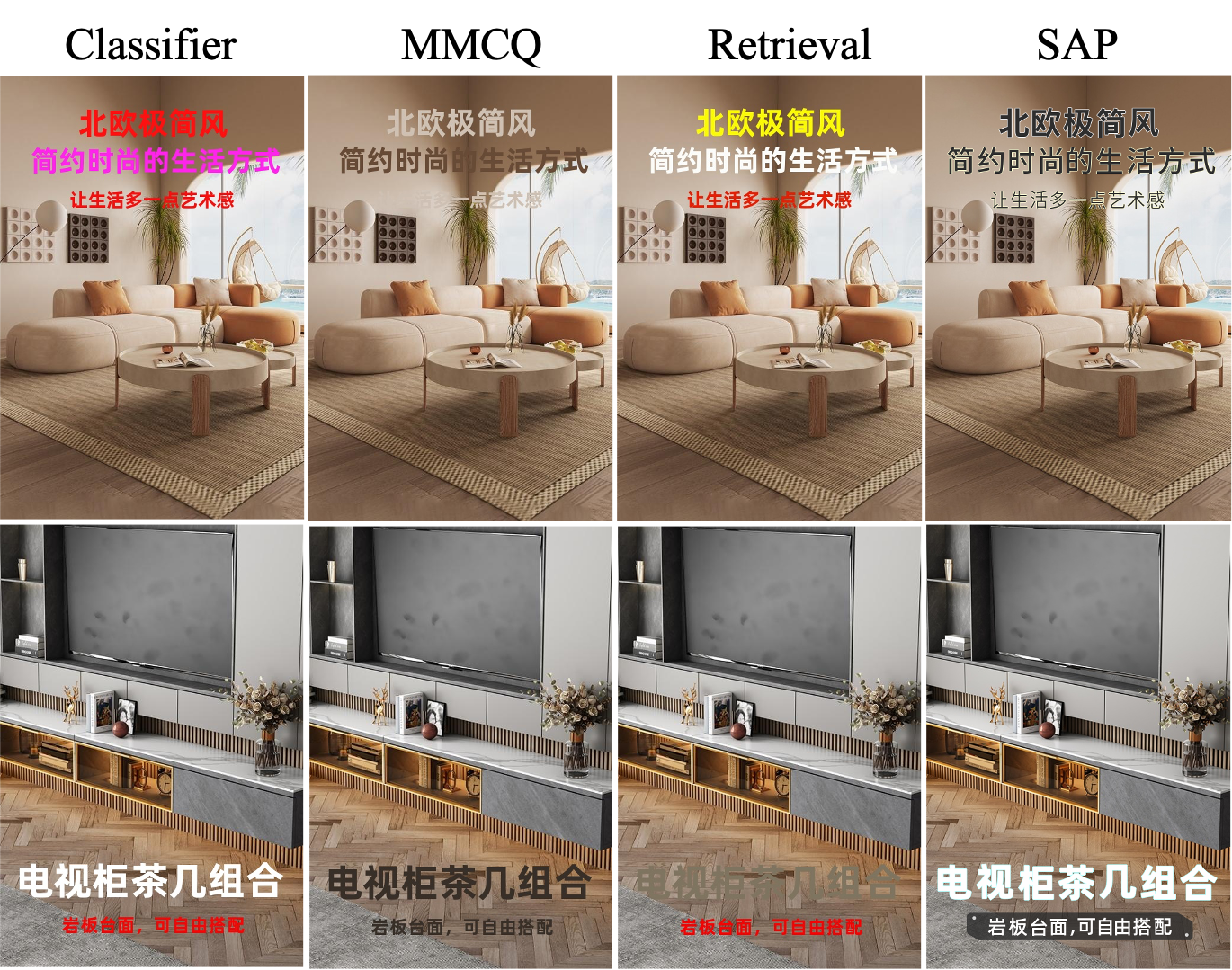}}
  \vspace{-8pt}
  \caption{Our SAP compared with Classifer~\cite{fontweb}, MMCQ~\cite{DBLP:conf/chi/VaddamanuAGSC22}, and Retrieval method for color prediction.}
  \vspace{-5pt}
  \label{fig:compare_baseline}
\end{figure}

Fig.~\ref{fig:compare_baseline} shows a visual comparison between SAP and other methods. As can be seen, the \textbf{Classifier} approach is prone to predict the color with the highest frequency, such as red, due to the problem of imbalanced color categories in the training data. The method of using \textbf{MMCQ} to extract the main colors of an image and increase the contrast of text color can lead to text that is unreadable. Moreover, none of the three competing methods take into account the correlation between colors for multiple sentences, resulting in inconsistent color combinations. In contrast, SAP can generate readable results with harmonious color combinations between different graphic elements. It can be seen that adding strokes and underlay further improves the readability of the tagline and the overall aesthetics.

\subsubsection{\textbf{Quantitative Evaluation}}
We assessed SAP, MMCQ, Retrieval, and Classifier on the test dataset using the D-ab \& D-light metric. As shown in the Table \ref{tab:sap_quali}, in consideration of the overall LAB space, SAP's predicted color is the most accurate. It is worth noting that SAP can estimate the dominant color, stroke color, gradient color, and font together, while the three comparison methods can only estimate the dominant color.

\begin{table}[h]
\vspace{-5pt}
\centering
\caption{Quantitative comparison results of SAP.}
\vspace{-10pt}
\label{tab:sap_quali}
\begin{tabular}{lcccc}
\toprule
Metric & MMCQ & Classifier & Retrieval & SAP \\
\midrule
D-ab$\downarrow$ & \textbf{22.08} & 40.72 & 33.40 & \underline{22.19} \\
D-light$\downarrow$ & 4.76 & 2.87 & \underline{2.65} & \textbf{2.33} \\
\bottomrule
\vspace{-5pt}
\end{tabular}
\end{table}

\vspace{-10pt}
\subsection{Ablation Studies}

We make two ablation studies to analyze the performance of SAP and gain insights into poster image design. Three metrics are used: a) dominant, gradient, and stroke color prediction error in the AB dimension of the LAB color space (D-ab, G-ab, and S-ab, respectively, calculated using mean squared error); b) dominant, gradient, and stroke color prediction error in the lightness dimension of the LAB color space (D-light, G-light, and S-light, respectively, calculated using mean absolute error); c) prediction accuracy of whether design elements possess gradient or stroke (G/S).

\begin{figure}[ht!]
  \centering
  \vspace{-5pt}
  \includegraphics[width=0.9\linewidth]{{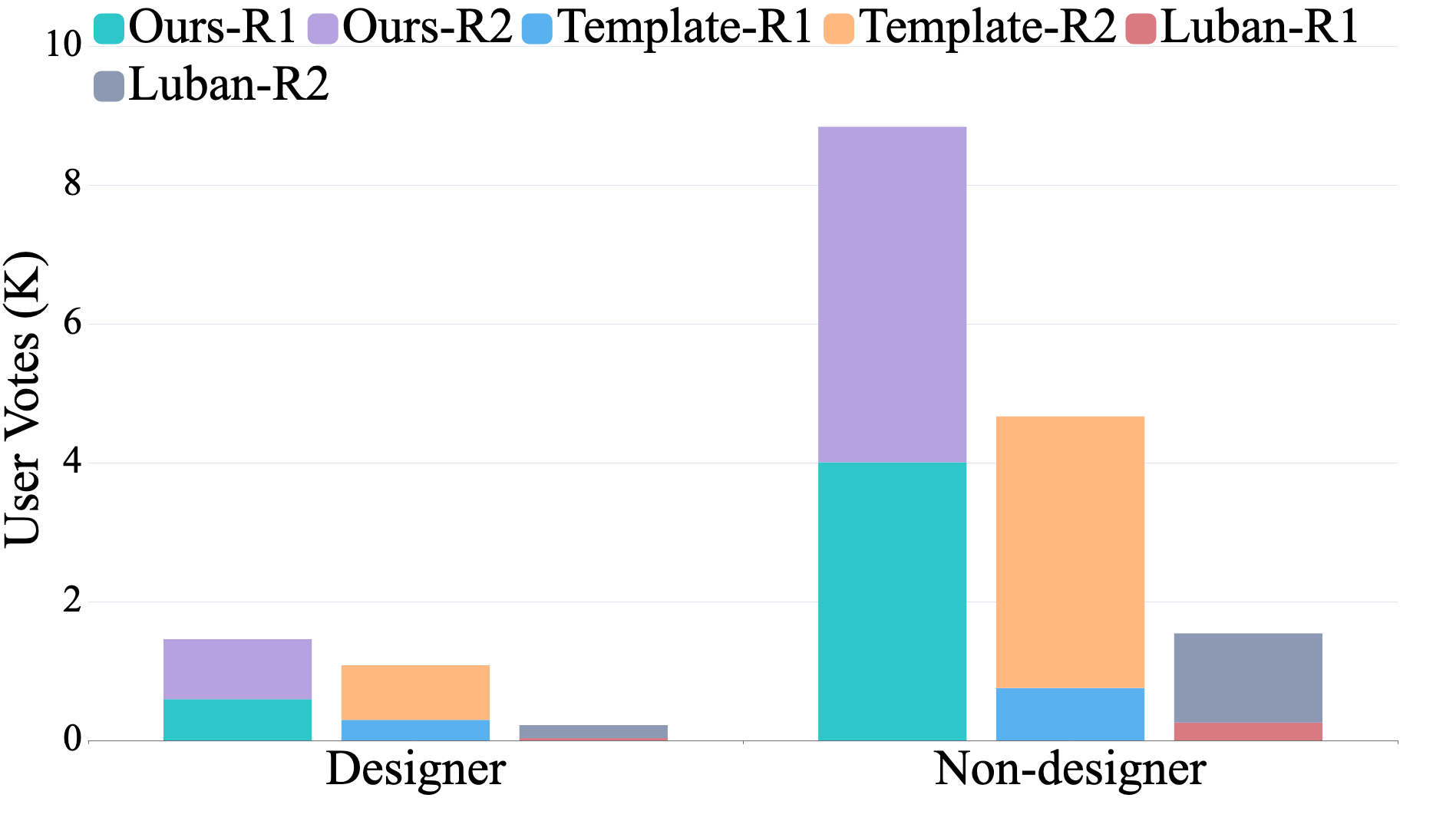}}
  \vspace{-10pt}
  \caption{Overall user survey results on R1 and R2 votes for different methods.}
  \label{fig:user_study}
  
\end{figure}

\begin{figure*}[htb!]
  \vspace{-10pt}
  \includegraphics[width=0.9\textwidth]{{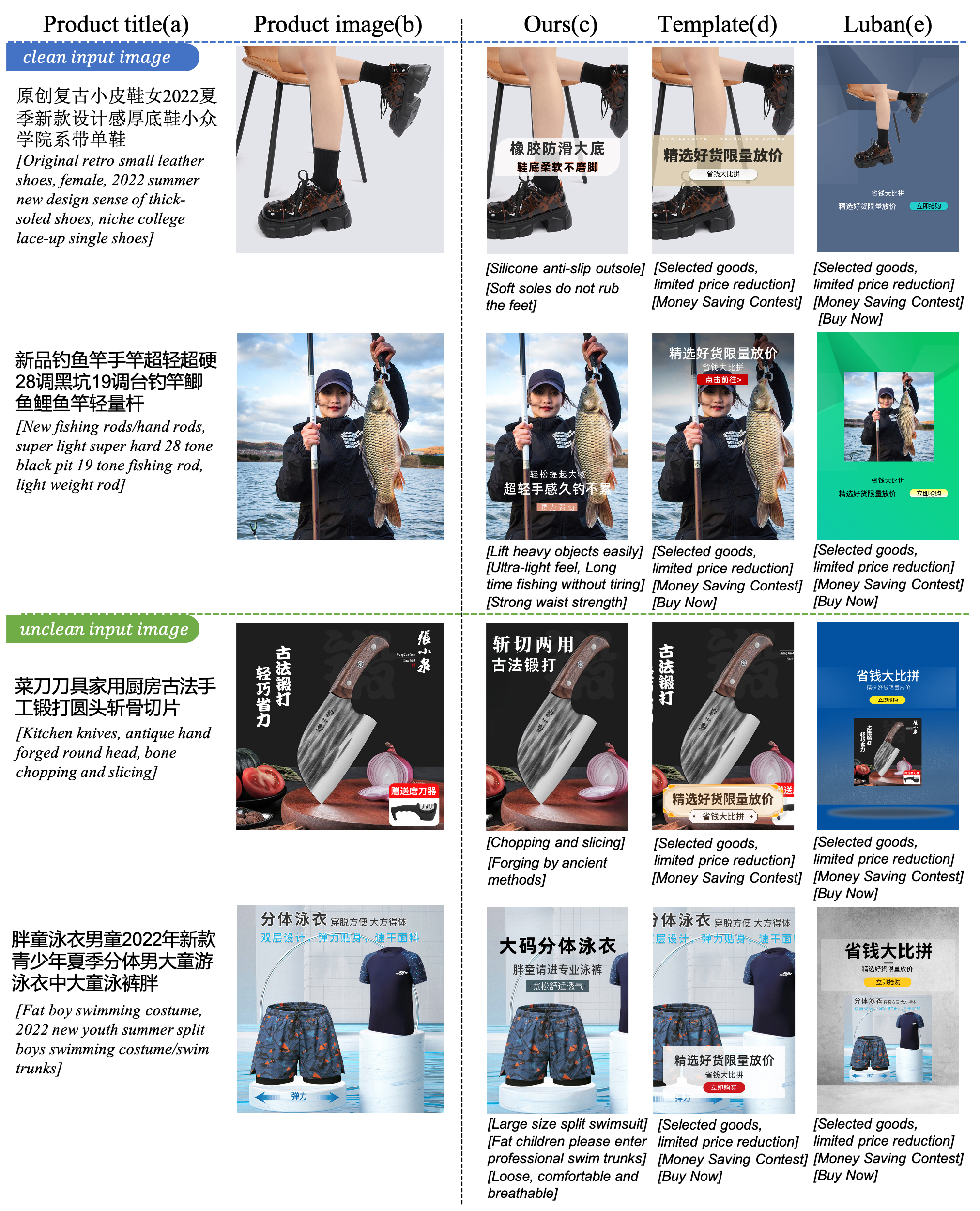}}
  \vspace{-10pt}
  \caption{Qualitative comparison of posters generated by various methods: (a) User-input Chinese product titles (descriptions); (b) User-uploaded images; (c, d, e) Posters produced by each of the three methods. "Unclean image" means there are no graphic design elements on the original input image, and vice versa.}
  \label{fig:comparison}
  \vspace{-10pt}
\end{figure*}

\begin{figure}[ht!]
  \centering
  \includegraphics[width=0.9\linewidth]{{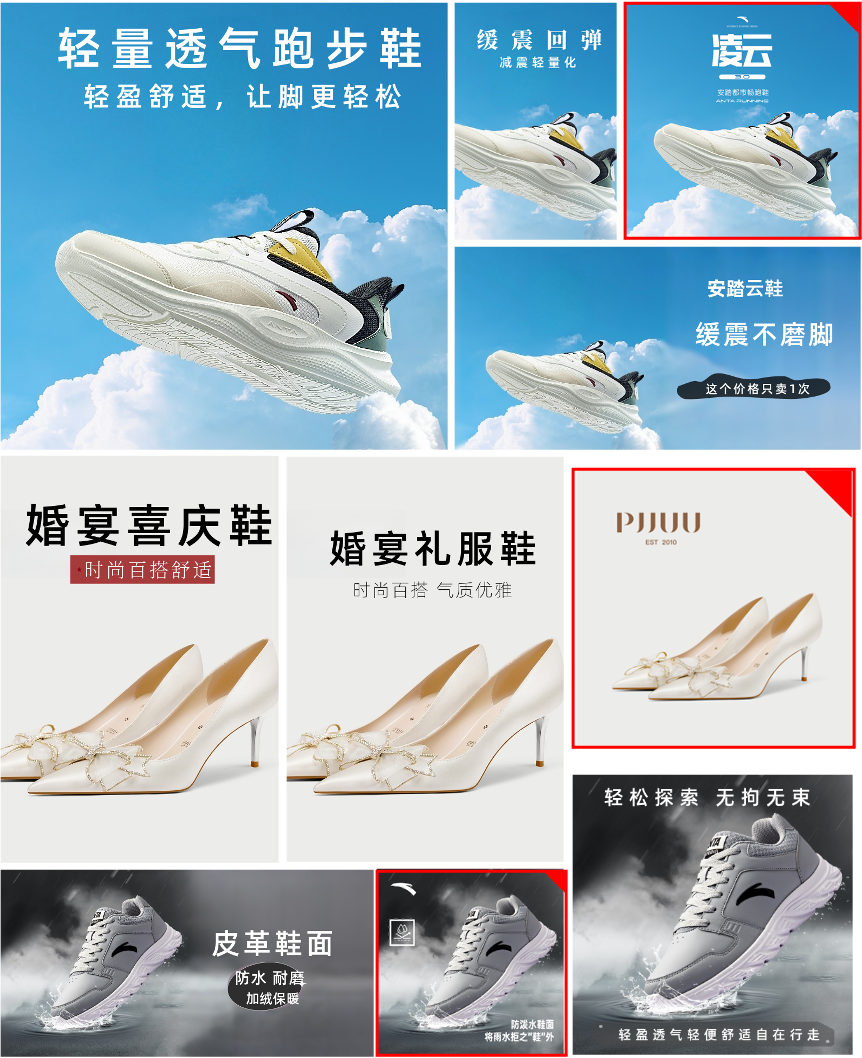}}
  \caption{Generated posters by AutoPoster in arbitrary sizes. The input product images are highlighted with \textcolor{red}{red box}.}
  \label{fig:multi_size}
  \vspace{-10pt}
\end{figure}

\vspace{-10pt}
\paragraph{\textbf{Single-task vs. multi-task learning.}}
To verify the effectiveness of joint optimization, we conduct four comparative experiments: 1) $Random$; 2) $D\text{-}color$: a $SAP$ model only predicting the dominant color; 3) $D\text{-}color+S\text{-}color$: a $SAP$ model predicting both the dominant and stroke color; 4) $D\text{-}color+G\text{-}color$: a $SAP$ model predicting the dominant and gradient color. Ours simultaneously predicts all three color attributes (dominant, stroke, and gradient color). The results are presented in Table \ref{tab:ablationstudy}. As the number of prediction tasks (attributes) increases, the performance improves. This suggests a correlation between the different color attributes in design, and learning one task can enhance the learning of other tasks.

\vspace{-3pt}
\paragraph{\textbf{Autoregressive vs. non-autoregressive manner}}
When designing the poster, the designer can follow two approaches: a) Decide on the overall layout first and then choose the visual style, considering the relationship between elements. b) Set the visual attributes of the most important element first and design the rest accordingly. These lead to two decoding methods: $NAR$ (non-autoregressive) and $AR$ (autoregressive). For a given product image $I$ and layout $E$, the probability of predicted attributes can be determined as follows:

\vspace{-10pt}
\begin{equation}
\begin{aligned}
& NAR: P_\theta(a_1, \dots, a_N | I, E) = \prod_{t=1}^{N} P_\theta(a_t | I, E)    \\
& AR: P_\theta(a_1, \dots, a_N | I, E) = \prod_{t=1}^{N} P_\theta(a_t | I, E, a_1, \dots , a_{t-1})
\end{aligned}
\end{equation}

In this equation, $a_i$ represents the style attributes of element $i$, and $\theta$ denotes the model parameters. The comparison of the last two rows in Table~\ref{tab:ablationstudy} demonstrates that $NAR$ outperforms $AR$ across all metrics. This suggests that, unlike machine translation (where $NAR$ typically yields lower translation accuracy~\cite{DBLP:journals/corr/abs-2204-09269}), the $NAR$ pattern is more consistent with poster design.

\subsection{Compare AutoPoster with Other Methods}

\vspace{-3pt}
\subsubsection{\textbf{User Study}}
To assess the visual quality of posters, we conducted a survey using a larger-scale questionnaire with settings similar to subsection 4.2. We take the 46 product images as input for test and use AutoPoster to generate posters (referred to as \textit{ours}) without further editing. To ensure a fair comparison, we use the \textit{Luban}\footnote{https://luban.aliyun.com} online system without subsequent editing. Additionally, we recruit designers to create generic poster templates. For each product image, we select a random template and replace the background image to obtain a poster, which we refer to as "\textit{Template}". Each product image corresponds to a question, with each question presenting three generated posters. It is worth noting that \textit{Luban} and \textit{Template} require manually inputting suitable taglines, so we use the default ones. To ensure fairness, participants are asked to make choices only based on visual quality (readability and aesthetics). A partial comparison of the produced advertising posters is shown in Figure~\ref{fig:comparison}.

In total, 129 participants completed the questionnaire, including 20 professional designers and 109 non-designers. As a result, 5,934 votes were collected. As shown in Fig.~\ref{fig:user_study}, the results demonstrate that our method outperforms the other methods in terms of both the $R1$ (78\%) and $R1+R2$ (58\%) metrics without
bells and whistles. 

\vspace{-5pt}
\subsubsection{\textbf{Online A/B Test}}
We conducted an online A/B test in a real e-commerce advertising scenario, randomly dividing online traffic users into two groups. One group viewed ads generated by AutoPoster, while the other group viewed ads created using the Template method. We did not use the \textit{Luban} method in this test due to inferior qualitative and quantitative results compared to the \textit{Template} method, which provided a stronger baseline. 
We tested 80,000 products, collecting around 30 million PV(page views) for each group. The AutoPoster group had a higher CTR (click-through rate) and RPM (Revenue Per Mille) increase of 8.24\% and 8.25\%, respectively, compared to the \textit{Template} group. These results demonstrate the effectiveness of our method in real industrial applications.

\subsection{Poster Generation with Various Sizes}
Our system can generate advertising posters of any size from a single input image, thanks to the proposed image cleaning and retargeting module. Each poster preserves the product's main features, and the graphics position\&scale adjusts accordingly, as shown in Figure~\ref{fig:multi_size}. Unlike cropping-only retargeting methods, our approach can expand the non-salient region of the original image, creating posters of various sizes. Linearly scaling design elements between posters of different sizes can result in disharmony if the aspect ratio significantly varies, such as from 2:3 to 3:2. However, AutoPoster can perceive these variations in all four design stages, creating posters adaptively for different target sizes.

\section{Conclusion and Future Work}
In this paper, we introduce AutoPoster, an innovative four-step approach for generating advertising posters that only requires product images, information, and user-specified poster dimensions to create complete posters. Additionally, we propose a novel Style Attribute Predictor that jointly learns multi-attribute prediction and an image retargeting technique that combines image cleaning to improve diversity and aesthetics in the generated posters. We also provide a large-scale dataset for poster generation. Our experiments demonstrate that AutoPoster and SAP produce posters with superior visual aesthetics compared to existing methods. In future work, we will extend this framework to encompass more forms of information representation design and explore variable font generation.


\clearpage
\bibliographystyle{ACM-Reference-Format}
\bibliography{acmart}

\appendix

\clearpage
\section{What Does Style Attribute Predictor Learn?}

\begin{figure}[h]
  \centering
  \includegraphics[width=0.5\textwidth]{{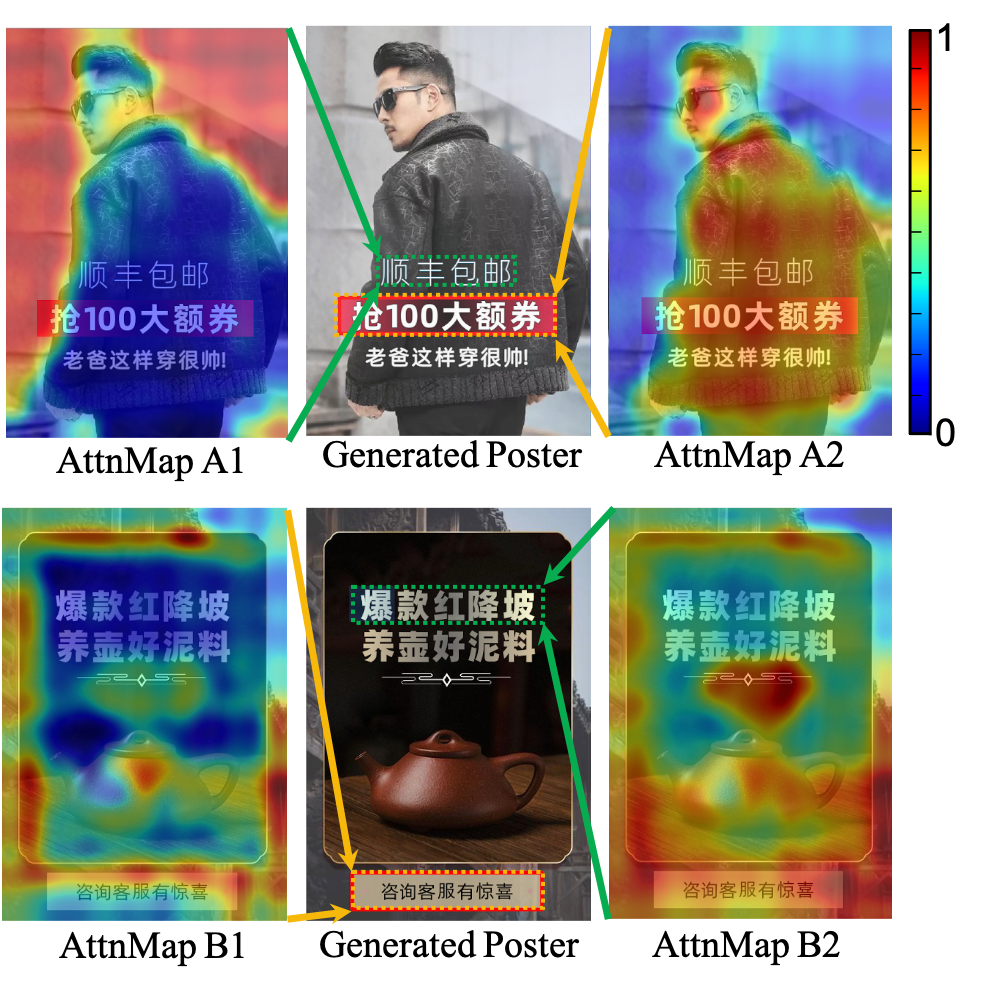}}
  \caption{
  Visualization of the attention map for SAP decoder. Each poster displays an attention map corresponding to a tagline element(\textcolor{green}{in green line}) and an underlay element(\textcolor{yellow}{in yellow line}). The heat map has been normalized to a range of 0 to 1.}
  \label{fig:attn_map}
\end{figure}

In this section, we attempt to explore the abstract and experiential nature of reasonable color matching between design elements for human designers. In order to gain insights into what the style predictor model learns, we visualize the last cross-attention layer of the style predictor decoder as shown in Fig.~\ref{fig:attn_map}. AttnMap B2 in Fig.\ref{fig:attn_map} illustrates that $SAP$ has captured the color change of the kettle from black to brown, explaining the prediction of the color gradient from bright yellow to black. An interesting observation was made that despite the lack of any supervised signal about image content, the model implicitly learned to locate and distinguish between the foreground (AttnMap A2 in Fig.~\ref{fig:attn_map}) and background (AttnMap A1 in Fig.~\ref{fig:attn_map}) regions of the image. Based on these observations, we conclude that the $SAP$ model has the ability to learn, to some extent, the pattern of color matching based on the appearance of the product subject and the overall color tone style of the image.

\section{Identification of Font Typography}

\begin{figure}[h]
  \centering
  \includegraphics[width=1.0\linewidth]{{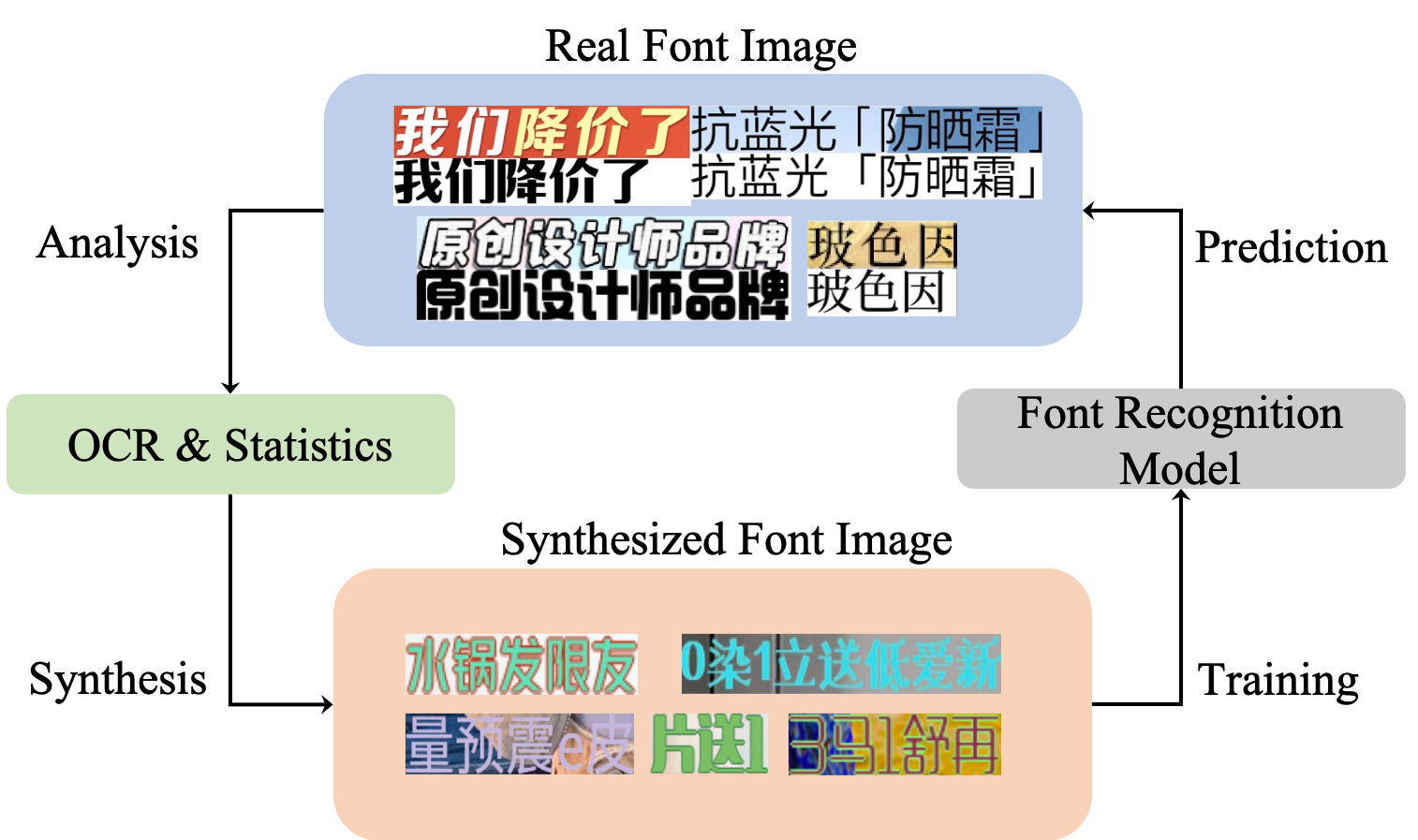}}
  \caption{Font image synthesis and font recognition workflow. For the images displayed in the "Real Font Image", the odd rows are the real font image and the even rows are the font recognized by the model.}
  \label{fig:font_id}
\end{figure}

To train the style predictor, it is necessary to have labeled attributes of graphic elements as supervisory signals. However, when dealing with taglines on rasterized images, manually identifying the font typography is challenging and remains an open-set problem. To address this issue, we propose a self-supervised training strategy for font typography recognition. 

The pipeline can be seen in Fig.~\ref{fig:font_id}. Initially, manually designed advertising posters are collected, and Optical Character Recognition (OCR) is employed to recognize the tagline contents in the poster images. Statistical analysis is then conducted on the Chinese character frequency, text size, and tagline length. Subsequently, taglines are rendered on randomly cropped clean background images based on these statistical findings. In the next step, a model is trained using 100k synthesized font images $I_{syn}$ as the model input, with the ground truth being the font typographies $f$ used for rendering. The backbone of this model is ResNet-50. It should be noted that the model only recognizes font typographies presented in the synthesized images. For font typographies outside this set, they may be recognized as a similar known one. The final trained model achieves an accuracy rate of 81.8\% on a test set containing 5,000 images.

\begin{figure*}[h]
  \centering
  \includegraphics[width=0.9\linewidth]{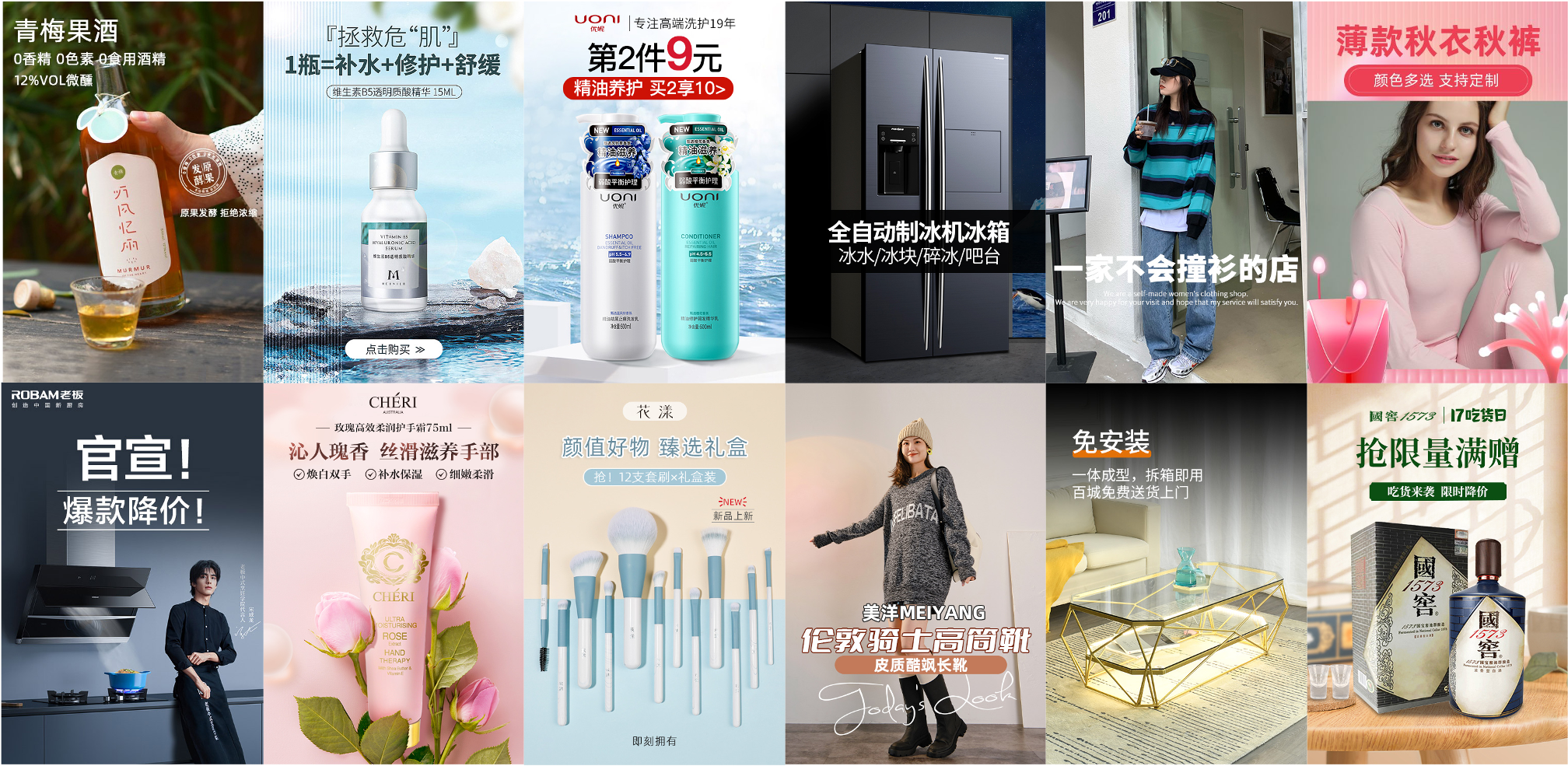}
  \caption{Visualization of a subset of posters extracted from the constructed dataset.}
  \label{dataset}
\end{figure*}

\begin{figure*}[h]
  \centering
  \includegraphics[width=0.9\linewidth]{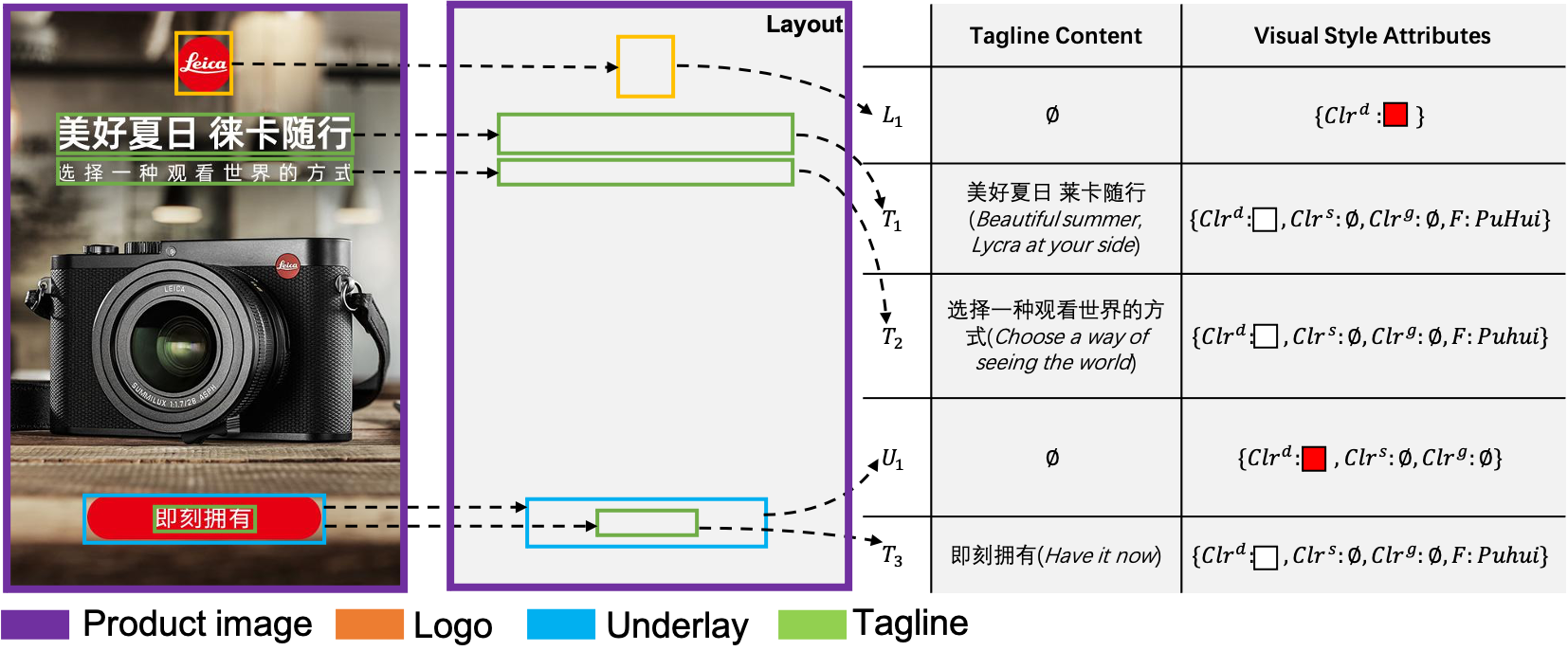}
  \caption{Annotation of an advertising poster. On the left is a poster with Chinese context. The layout of graphic elements on this poster is shown in the middle. And the right part lists the tagline content and visual style attributes of each design element. $PuHui$ is a kind of Chinese font typography.}
  \label{designspace}
\end{figure*}

\section{Poster and annotation example.}
Fig.~\ref{dataset} shows some posters of the constructed dataset mentioned in the paper. Additionally, Fig.~\ref{designspace} provides an example of the poster annotations. The annotation encompasses three key aspects: layout, tagline content, and visual style attribute.

\section{Limitations}
Our method has two main limitations. Firstly, the available underlays are currently limited in number and exhibit simplistic styles. This issue may be released by constructing a larger and more diverse library of underlays or introducing supplementary embellishments. Secondly, our method can only predict the color and several font typographies of taglines, lacking the ability to generate variable or artistic fonts at the pixel level (e.g.~\cite{Wang_2022_CVPR}).


\end{sloppypar}
\end{document}